# Evaluation of Taxonomic and Neural Embedding Methods for Calculating Semantic Similarity

Dongqiang Yang[*] and Yanqin Yin

School of Computer Science and Technology
Shandong Jianzhu University, China
[*]Correspondence: ydq@sdjzu.edu.cn

Abstract

Modelling semantic similarity plays a fundamental role in lexical semantic applications. A natural way of calculating semantic similarity is to access hand-crafted semantic networks, but similarity prediction can also be anticipated in a distributional vector space. Similarity calculation continues to be a challenging task, even with the latest breakthroughs in deep neural language models. We first examined popular methodologies in measuring taxonomic similarity, including edge-counting that solely employs semantic relations in a taxonomy, as well as the complex methods that estimate concept specificity. We further extrapolated three weighting factors in modelling taxonomic similarity. To study the distinct mechanisms between taxonomic and distributional similarity measures, we ran head-to-head comparisons of each measure with human similarity judgements from the perspectives of word frequency, polysemy degree, and similarity intensity. Our findings suggest that without fine-tuning the uniform distance, taxonomic similarity measures can depend on the shortest path length as a prime factor to predict semantic similarity; in contrast to distributional semantics, edge-counting is free from sense distribution bias in use and can measure word similarity both literally and metaphorically; the synergy of retrofitting neural embeddings with concept relations in similarity prediction may indicate a new trend to leverage knowledge bases on transfer learning. It appears that a large gap still exists on computing semantic similarity among different ranges of word frequency, polysemous degree, and similarity intensity.

## 1. Introduction

Similarity plays a fundamental role in Artificial Intelligence (AI) and human cognition processes such as reasoning, memorizing, and categorization. Specifically, in natural language processing (NLP), lexical semantic similarity can decrease a model's generalization errors on unseen data and facilitate interpreting word meanings in contexts, among others. Calculating such semantic similarity usually employs two resources: hand-crafted semantic networks contained within knowledge bases and lexical co-occurrence patterns collected in contexts, which brings forth two respective tasks of measuring taxonomic and distributional similarity. Taxonomic similarity is more associated with semantic similarity rather than relatedness, whereas distributional similarity hardly distinguishes semantic similarity from relatedness (Hill et al. 2015, Lê and Fokkens 2015). The fusion of taxonomic and distributional similarity in a unified model appears to be a perfect solution to harvesting their advantages on similarity calculation. For example, Banjade et al. (2015) trained an ensemble of taxonomic similarity on WordNet (Pedersen et al. 2004) and distributional similarity on both Wikipedia (Gabrilovich and Markovitch 2007) and neural network embeddings (NNEs) (Huang et al. 2012, Mikolov et al. 2013b, Pennington et al. 2014) using Super Vector Regression. On the assumption that semantically linked words should stay closer in a distributional space, Wieting et al. (2015) fine-tuned a pre-trained neural network embeddings (Mikolov et al. 2013b) through retrofitting word pairs extracted from the Paraphrase Database (Ganitkevitch et al. 2013). Faruqui et al.

(2015) employed concept relationships in WordNet (Miller et al. 1990, Fellbaum 1998), FrameNet (Baker et al. 1998), and the Paraphrase Database (Ganitkevitch et al. 2013) to enhance NNEs.

Despite the latest advances on state-of-the-art neural similarity models (Faruqui et al. 2015, Wieting et al. 2015, Mrkšić et al. 2017, Shi et al. 2019), quantifying to what extent words are kin to each other, and even identifying the underlying relationships between words, is still a challenging task. Open questions remain about the effectiveness of weighting factors such as path length and concept specificity in modelling taxonomic similarity. An unaddressed problem in the literature is a lack of comparisons of the effectiveness of multiple weighting factors in yielding taxonomic similarity. Moreover, with the advent of neural language models (NLMs), further investigations on taxonomic and distributional similarity measures are also needed to pinpoint their strengths and weaknesses in similarity prediction.

To address the aforementioned problem, we first study to what extent different weighting factors can function in modelling taxonomic similarity; we then contrast taxonomic similarity with distributional similarity on the benchmark datasets. The main contributions of this paper are threefold. First, after analysing the popular taxonomic similarity measures, we generalized three weighting factors in modelling taxonomic similarity, among which the shortest path length in edge-counting can hold a prime role in similarity prediction. The advantage of edge-counting clearly lies in that concept similarity can work on all comparisons of concept relationships, irrespective of conceptual metaphor, literalness, or usage statistics. Second, using the gold standard of human similarity judgements, we demonstrate that edge-counting can outperform the unified and contextualized neural embeddings, and the simple edge-counting can effectively yield semantic similarity without the involvement of conditioning factors on the uniform distance. Third, further analysis on the different levels of word frequency, polysemy degree, and similarity intensity shows that during their self-supervised learning process, neural embeddings may be susceptible to word usage variations; the unbalanced domains and concept coverage in hand-crafted semantic networks may impose a constraint on edge-counting. To combine taxonomic and distributional similarity efficiently, retrofitting neural embeddings with concept relationships may not only improve semantic similarity prediction but also provide a better way of transfer learning for downstream applications in NLP.

Note that as the main goal of this paper is to investigate semantic similarity measures using NNEs and paradigmatic relations in taxonomies, we distinguish the use of semantic similarity and relatedness in the paper. Pure semantic similarity presupposes the overlapping of the semantic features of concepts in semantic memory (Quillian 1968) and their spreading-activation in human cognition (Collins and Loftus 1975), whereas pure relatedness reflects word co-occurrences so that one word can stimulate the presence of other words, whereby the semantic features of the stimulus can be carried. Their distinction is not absolute, such as *mother* and *father*; and *coffee* and *tea*, since these words can be both semantically and associatively related. Moreover, the paradigmatic aspect of word meanings emphasizes inter-textual substitution in a paradigm, whereas the syntagmatic aspect is concerned with intra-textual co-occurrence. Both are two facets to account for word meanings, and semantic similarity is more like a particular case of semantic relatedness (Budanitsky and Hirst 2006). Hence semantic relatedness is composed of both word similarity with approximation intensity and word association with association magnitude.

The rest of the paper is structured as follows: Section 2 briefly introduces related work on evaluating taxonomic and distributional similarity; Section 3 examines different weighting factors in calculating taxonomic similarity, mainly investigating the hypothesis of concept specificity in fine-tuning the uniform distance; Section 4 outlines major neural embeddings models for calculating distributional similarity; Section 5 conducts an in-depth evaluation of taxonomic and distributional similarity on the benchmark datasets in different ranges of word frequency, polysemy degree, and similarity intensity; Section 6 further discusses major discrepancies among the similarity measures through analysing their mutual correlation patterns; Section 7 concludes with several observations and future work.

## 2. Related work

There is a considerable amount of literature on evaluating taxonomic similarity measures. Apart from using human similarity ratings (Agirre et al. 2009, Hill et al. 2015) to validate the measures, many surveys have focused on evaluating them using downstream applications. For example, McCarthy et al. (2004a, 2004b) investigated the methods implemented in the similarity package (Pedersen et al. 2004) on how to acquire the predominant (first) sense automatically in WordNet from raw corpora; Pedersen et al. (2005) systematically examined the taxonomic similarity methods in their similarity package through word sense disambiguation (WSD); Budanitsky and Hirst (2006) evaluated some taxonomic similarity measures, partly contained in the package, through malapropism recognition. They all found that JCN (Jiang and Conrath 1997), employing both WordNet taxonomy and corpus statistics, outperformed most models in the package.

The WordNet-based similarity measures (Pedersen et al. 2004) were also surveyed for studying their adaptability on different domains of lexical knowledge bases (LKBs) such as Roget (McHale 1998, Jarmasz and Szpakowicz 2003) and the Gene Ontology (GO) (Pedersen et al. 2007, Guzzi et al. 2011). In calculating semantic similarity and relatedness, Strube and Ponzetto (2006) investigated the measures respectively with WordNet and Wikipedia. Their investigation indicated that Wikipedia was less helpful than WordNet in semantic similarity computation. However, it outperformed WordNet when deriving semantic relatedness, which could be attributed to more abundant concept relationships encoded in Wikipedia than in WordNet. Instead of counting concept links in the category tree of Wikipedia (Strube and Ponzetto 2006), Gabrilovich and Markovitch (2007) proposed the Explicit Semantic Analysis (ESA) to incorporate concept frequency statistics from Wikipedia in semantic relatedness calculation. They showed that ESA was superior to other Wikipedia or WordNet-based similarity methods. Zesch and Gurevych (2010) systematically reviewed the measures on the folksonomies such as Wikipedia and Wiktionary. They found that these collectively constructed resources were not superior to the well-compiled LKBs in deriving semantic relatedness.

Apart from extensive investigations on distributional similarity measures (Curran 2003, Dinu 2012, Panchenko 2013), a growing body of literature has analysed various basis elements in vector space models (Turney and Pantel 2010). These elements form the dimensionality of a semantic space either through syntactically conditioned (syntactic dependencies) (Curran 2003, Weeds 2003, Yang and Powers 2010, Panchenko 2013) or unconditioned (a bag of words) (Bullinaria and Levy 2006, Sahlgren 2006) co-occurrence settings. Moreover, Padó and Lapata (2007) compared the two settings with a series of evaluations on semantic priming in cognitive science, multiple-choice synonym questions, and WSD. Mohammad and Hirst (2012) further conducted a systematic survey on many taxonomic and distributional similarity models and identified pros and cons of yielding semantic similarity and relatedness through semantic networks and co-occurrent lexical patterns, respectively. They concluded that both models should be evaluated in terms of contrasting with human similarity judgements to better disclose their distinctive properties so that they can function complementally to achieve human-level similarity prediction.

Most current surveys on knowledge-based similarity measures have focused either on how to select semantic relatedness measures using knowledge resources, computational methods, and evaluation strategies (Feng et al. 2017) or on how to adapt the WordNet-based taxonomic similarity into other domains such as biomedical ontologies (Zhang et al. 2013) and a multilingual BabelNet of integrating both WordNet and Wikipedia (Navigli and Ponzetto 2012a, b). Note that Harispe et al. (2015) proposed a comprehensive survey on both knowledge-based and counts-based distributional similarity measures. Although some work on the potential of prediction-based neural embeddings (Huang et al. 2012, Mikolov et al. 2013b, Pennington et al. 2014) on yielding semantic similarity has been carried out (Hill et al. 2015, Gerz et al. 2016), there are still lack of extensive surveys on the difference between taxonomic structure and NNEs in calculating semantic similarity. Especially with the recent developments on the contextualized NNEs (Devlin et al. 2018, Howard and Ruder 2018, Peters et al. 2018a, Radford et al. 2018) and NNEs retrofitted with semantic relationships (Faruqui et al. 2015, Wieting et al. 2015, Mrkšić et al. 2016, Mrkšić et al. 2017, Ponti et al. 2018, Shi et al. 2019), further research should be undertaken to differentiate the NNEs-derived distributional similarity from taxonomic similarity. Overall, the aim of our work is to extend current knowledge of both taxonomic similarity in WordNet and distributional similarity in NNEs.

## 3. Universal edge-counting in a taxonomy

Edge-counting or shortest path methods on taxonomic similarity assume that the shortest distance entails the most substantial similarity between concepts. It can be traced back to Quillian's semantic memory model (Quillian 1967, Collins and Quillian 1969): concept nodes are located within a hierarchical network; the number of links between the nodes can correspondingly predict their semantic distance.

In Figure 1, the root node is the top superordinate in a taxonomy, and the direct connection between two immediate concepts is called a link or hop that stands for concept relationship. The length of a path or route from concept 1 to concept 2 is equal to the number of links when traversing between them. If all concept nodes are situated in the leaves of the hierarchy, the shortest path length (*spl*) or shortest distance (*sd*) along the links (edges) between them can travel through their nearest common node (*ncn*). To depict the specificity of concepts in a taxonomy, we define the depth (*dep*) of a concept which is *spl* from the concept to the root; and we define the local density (*den*) of a concept as the number of its descendants (the number of links leaving from a concept) or the number of its siblings (the number of links from leaving its parent node).

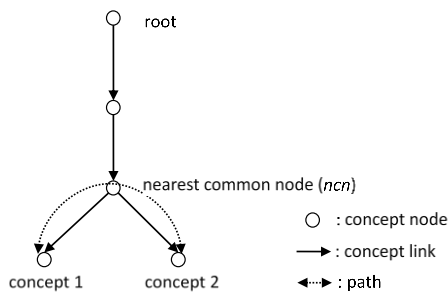

Figure 1: A diagram of edge-counting in a taxonomy.

We first symbolize the $m^{th}$ sense of word $i$ as $w_{i,m}$, and the $n^{th}$ sense of word $j$ as $w_{j,n}$. The range of $m$ is from 1 to $|w_{i,m}|$ that is the total number of senses of $w_i$ in WordNet (Miller 1995, Fellbaum 1998). So too the range of $n$ is from 1 to $|w_{j,n}|$. Given that the shortest distance between concepts: $w_{i,m}$ and $w_{j,n}$, is the sum of the shortest distances between each concept and *ncn* (the nearest superordinate in a IS-A hierarchy), $sd(w_{i,m}, w_{j,n})$ can be defined as: $sd(w_{i,m}, w_{j,n}) =$

$dep(w_{i,m}) + dep(w_{j,n}) - 2 \times dep(ncn(w_{i,m}, w_{j,n}))$ on condition that *ncn* can be traced in the hierarchy.

### 3.1. Simple edge-counting

Rada et al. (1989) defined the *Distance* model to explain concept distance as the *minimum number of edges separating a and b* given that *a* and *b* represent the concepts in an IS-A hierarchy of MeSH. They argued that the *Distance* model satisfied the three axioms of the metric in a geometric model (Torgerson 1965, Goldstone 1994), namely minimality, symmetry, and triangular inequality. In the following sections, we equate the *Distance* model with the simple edge-counting.

### 3.2. A variant of the simple edge-counting

Leacock and Chodorow (1994, 1998) proposed a similarity model based on the taxonomy of WordNet to optimise the local context classifier in disambiguating word senses:

$$Sim(w_i, w_j) = Max\left[-log\frac{dist(w_{i,m}, w_{j,n})}{2*D}\right]$$

where $dist(w_{i,m}, w_{j,n})$ is the distance or path length between concepts $w_{i,m}$ and $w_{j,n}$, and *D* is a constant, equal to the maximum depth (16) in the taxonomy of WordNet. Note that they valued the distance between concepts on the number of interior nodes along paths plus one in their original definition of the model, no different from the simple edge-counting. They also defined the similarity of two words as the maximum of all pairwise concept similarities. Since maximizing similarity scores in the model corresponds to finding the shortest path length, the model is equivalent to an adaptation of the *Distance* model in WordNet after removing the logarithmic normalization.

### 3.3. Concept specificity

In searching for the shortest path in semantic networks, Rada et al. (1989) claimed that the premise of uniform distance in edge-counting should hold differently under different relationships and substructures. Concept similarity is supposed closer if their *ncn* sits in a deeper or denser subpart of semantic networks where concept connotation is then more specific. Concept depth and density were two widely employed factors in estimating concept specificity (Sussna 1993). In contrast to using word frequency statistics to formulate concept specificity, e.g. information content (Resnik 1995), the following methods only employed taxonomy structures. We henceforth also named such methods as intrinsic information content, initially used by Seco et al. (2004) for their method of calculating a concept's specificity by counting the number of its hypernyms.

#### 3.3.1 Concept depth

Wu and Palmer (1994), in machine translation from English to Chinese, proposed to map verb senses into different semantic domains, and then to measure verbal sense similarity in each projected domain:

$$Sim(w_{i,m}, w_{j,n}) = \frac{2*dep(ncn(w_{i,m}, w_{j,n}))}{dep(w_{i,m}) + dep(w_{j,n})}$$

Since the shortest distance *sd* is equal to $dep(w_{i,m})+dep(w_{j,n})-2\times dep(ncn(w_{i,m}, w_{j,n}))$, verb similarity can be expressed as:

$$Sim(w_{i,m}, w_{j,n}) = 1 - \frac{sd(w_{i,m}, w_{j,n})}{dep(w_{i,m}) + dep(w_{j,n})}$$

where *sd* is normalized by the sum of the concept depths. We can view it as a depth-normalized variant of the *Distance* model, and the premise of the model contrasts with the uniform distance in the *Distance* model. If concept depth is an indicator of the specificity of the word senses for the same shortest distance, the concepts located in the lower part of a hierarchy would be more specific in meaning. They would be more akin to each other than the relatively general concepts in the upper part of the hierarchy, although sometimes the nuance in dissimilarity is hard to distinguish.

Motivated by the work of using concept distance to disambiguate word senses (Rada et al. 1989, Sussna 1993), Agirre et al. (1994) also proposed a varied concept depth model to calculate concept difference. They assumed that concept distance should be the shortest path length in the Intelligent Dictionary Help System (IDHS), a semantic network of sense frames connected by syntagmatic and paradigmatic relationships. Their concept distance or dissimilarity model can be expressed as:

$$dist(w_{i,m}, w_{j,n}) = \sum_{k=1}^{|sd(w_{i,m},w_{j,n})|} 1/dep(w_k)$$

where $|sd(w_{i,m},w_{j,n})|$ is the number of all concept nodes ($w_k$) along the shortest path between $w_{i,m}$ and $w_{j,n}$.

### 3.3.2 *Local concept density*

The same length of a link in a sub-hierarchy could have different weights that depended on the density of the sub-hierarchy. For example, without other relationships except for hypernyms, concept similarity in a local area with three hypernym links will be less than with two hypernym links. The higher the local density held, the lower similarity between the concepts was. Sussna (1993) postulated that some link types deserve two different weights, given that a link represents a two-way relationship (a bidirectional link). For example, an IS-A link constituting both hypernym and hyponym should have two different weights, as does a HAS-A link with both holonym and meronym. The omnidirectional relations of synonym and antonym were assigned the same weights. Concept distance in each direction is reciprocally proportional to semantic similarity, which can be expressed:

$$dist(w_{i,m} \rightarrow_r w_{j,n}) = max_r - \frac{max_r - min_r}{nrl(w_{i,m} \rightarrow_r)}$$

In this case, $min_r$ and $max_r$ were minimum and maximum weights for the relation *r*, which represented the connection strength between concepts. They were established with 0 for synonyms, 2.5 for antonyms, and between 1 and 2 for hyper/hyponyms and holo/meronyms. *nrl* stands for the number of relationship links. Note that $nrl(w_{l,m}\rightarrow_r)$ is the link counts starting from $w_{l,m}$ along the same *r*, which can act as a normalization factor, i.e. the *type specific fanout factor (tsf)*. In Sussna's work, this reduced the link strength, which leads to the concept distance increased with the growing number of links for relation *r*. Thus, *tsf* can also be assumed to be a local concept density.

Normalized by concept depth, concept distance on one link in the hierarchy of WordNet, uniformly equal to 1 in edge-counting, can be reformulated as:

$$dist(w_{i,m}, w_{j,n}) = \frac{dist(w_{i,m} \rightarrow_r w_{j,n}) + dist(w_{j,n} \rightarrow_{r^{-1}} w_{i,m})}{2 \times Max(dep(w_{i,m}), dep(w_{j,n}))}$$

where $r^{-1}$ is the inverse of *r*, and $Max(dep(w_{i,m}), dep(w_{j,n}))$ is the maximum of the depth of two nodes rather than the sum of the concept depths in the similarity model of Wu and Palmer

(1994). Thus, within the same shortest distance, specific concepts are closer than general ones. The semantic distance of two concepts is the sum of all the link distances along the shortest path between them.

### 3.3.3 *Path type*

Hirst and St-Onge (1997) established multi-level weights on different lexical relationships (e.g. IS-A and HAS-A) in the taxonomy of WordNet to construct lexical chains to resolve the problem of detection and correction of malapropisms. The weights ranged from high to low, which include:

- **Extra-strong**: the highest weight if two words were identical in morphological form.
- **Strong**: weaker than extra-strong when two words are synonymous or antonymous, or one word is a part of another word (a phrase).
- **Medium-strong**: assigned for the IS-A relationships. This can be further formulated as:

$Sim(w_i, w_j) = Max[Sim(w_{i,m}, w_{j,n})] = Max[C - dist(w_{i,m}, w_{j,n}) - K*dir(w_{i,m}, w_{j,n})]$

where C and K are constants, respectively 8 and 1 in their experiment, and $dir(w_{i,m}, w_{j,n})$ is the number of direction changes when searching the possible paths from $w_{i,m}$ and $w_{j,n}$ in WordNet, and the distance $dist(w_{i,m}, w_{j,n})$ is restricted from 2 to 5 links.

Hirst and St-Onge are the first to weight the link changes in the concept similarity model, although they did not fine-tune the relationship between distance and the link changes. Their method was extendable with more abundant relationships, i.e. paradigmatic and syntagmatic, accessible in Roget's Thesaurus (Morris and Hirst 1991).

### 3.3.4 *Hybrid utilization of concept specificity*

Apart from the path type (Hirst and St-Onge 1997) in quantifying conceptual relationships, Yang and Powers (2005, 2006) proposed a hybrid weighting scheme that took into account both the path type ($\alpha_t$) and the link type ($\beta_t$) for a relationship $t$ in improving their edge-counting similarity model. Supposing that the concept distance in a taxonomy is the least number of links between conceptual nodes, they defined the similarity of two concepts as:

$$Sim(w_{i,m}, w_{j,n}) = \alpha_t \times \beta_t^{sd(w_{i,m}, w_{j,n})-1}, \text{ iff } sd(w_{i,m}, w_{j,n}) \leq \gamma. \text{ Otherwise, } 0.$$

where $0 \leq Sim(w_{i,m}, w_{j,n}) \leq 1$ and

- $sd(w_{i,m}, w_{j,n})$: the shortest distance ($dist(w_{i,m}, w_{j,n})$) between concepts $w_{i,m}$ and $w_{j,n}$;
- $\gamma$: the depth limit factor, which is an arbitrary threshold on the distance introduced both for efficiency and to represent human cognitive limitations;
- $t$ = *hh* (hyper/hyponym), *hm* (holo/meronym), *sa* (syn/antonym), *id* (identical).

Apart from the three factors in the hybrid model to cope with taxonomic similarity on nouns, Yang and Powers (2006) also introduced the other three fallback factors: $\alpha_{der}$, $\alpha_{stm}$, and $\alpha_{gls}$ to assess it on verbs. Note that *der* (derived nouns), *stm* (stemming), and *gls* (gloss) stand for three extra relationships introduced for verbs. They found that even without the contributions of the three fallback factors, the verb similarity model relying on the shortest path length, the path type, and the link type, analogous to the noun model, still performed competitively well.

Note that we can treat the process of yielding word similarity between $w_i$ and $w_j$ as finding the shortest distance of their corresponding senses or the maximum of all concept similarity. This description follows the assumption in edge-counting that the shortest distance between

concepts indicates the highest similarity in semantics. Therefore, the hybrid model can be seen as:

$$\operatorname*{argmax}_{m,n} \left( \alpha_t \times \beta_t^{dist(w_{i,m}, w_{j,n})-1} \right) (0 < \alpha_t, \beta_t < 1)$$
$$= \operatorname*{argmax}_{m,n}(\log \alpha_t + (dist(w_{i,m}, w_{j,n}) - 1) \log \beta_t)$$
$$= \operatorname*{argmin}_{m,n}(-\log \alpha_t - (dist(w_{i,m}, w_{j,n}) - 1) \log \beta_t)$$
$$= \operatorname*{argmin}_{m,n}(dist(w_{i,m}, w_{j,n}) + (\log \alpha_t - \log \beta_t)/\log \beta_t)$$

After the logarithm operation, the hybrid model is similar to the edge-counting except that *α*, *β*, and *γ* are adjustable in different hierarchies and applications. It indicates that the shortest distance can be the dominant factor in the hybrid model without employing any normalization of the local concept density and concept depth in the IS-A and HAS-A hierarchies.

### 3.4. *Information content*

To fine-tune the uniform distance, we can rely on the intrinsic structures in LKBs such as concept depth and density in estimating concept specificity. Conversely, word or concept usage statistics acquired from corpora, not only indicating concept salience in language usages but also reflecting language development, can also serve for calculating concept specificity.

Resnik (1995) argued that the uniform distance of a link in WordNet posited by the simple edge-counting could not fully account for the semantic variability of a single link in practice. To avoid bias in the measurement of the uniform distance, Resnik proposed to model Information Content (IC) of *ncn*, i.e. minimum upper bound in a hierarchy, to augment the simple edge-counting, which is: $Sim(w_{i,m}, w_{j,n}) = IC(ncn(w_{i,m}, w_{j,n})) = -\log P(ncn(w_{i,m}, w_{j,n}))$. Here the shortest path length between $w_{i,m}$ and $w_{j,n}$ across their $ncn(w_{i,m}, w_{j,n})$ was substituted with the probability of $ncn(w_{i,m}, w_{j,n})$, which was calculated by the sum of the probabilities of all subordinate concepts under $ncn(w_{i,m}, w_{j,n})$ in 1 million words of the Brown Corpus of American English. Resnik's hybrid similarity model using both the simple edge-counting and frequency statistics resembles the similarity model proposed by Leacock and Chodorow (1994, 1998), which only relies on counting edges with the logarithm of the shortest distance. Note that the probability computation of *ncn* in information content is not concerned with the contribution of the word co-occurrences in accounting for word similarity; instead, it deals with the simple accumulation of the occurring events of a single word under *ncn*. Another shortcoming of information content is that *ncn* often serves for all the concepts under it in computing similarity. All the concepts are consequently similar to each other, even if they are situated on different levels in a taxonomy and are essentially different in the shortest distance under *ncn*.

In a hybrid similarity model, Jiang and Conrath (1997) explored the linear combination of information content (Resnik 1995) with path types, concept depths, and local concept densities that were explored in Sussna's work (1993). Contending that the links between concepts and their *ncn* in the IS-A hierarchy of WordNet are not satisfied with the requirement for uniform distance, they varied the link weight with the IC of the links rather than the concepts. They calculated the conditional probability of a concept given its *ncn*, in other words, the IC of *ncn* subtracting the IC of the concept, given that the probability of the *ncn* is not zero. The computation of the link weight in this way implies that a concept and its *ncn* have to co-occur in the corpora. Since they only employed IS-A links to find paths, Jiang and Conrath found that the path type factor imposed no impact on quantifying word similarity. The local density of *ncn*, standing for the number of the hyponyms in the subpart of the hierarchy between

*ncn* and two concepts, as well as the depth of *ncn*, should work together to complement the unbalanced concept distribution in the hierarchy. This is a similar idea to Sussna's concept distance model (1993). All of these factors were systematically merged into the function of the path weights between concepts and their *ncn*, and so concept distance is simply the sum of these path weights.

When cutting off the depth and local density weights in their similarity model, concept distance can be simplified as follows:

$$PW(w_{i,m}, w_{j,n}) = IC(w_{i,m}) + IC(w_{j,n}) - 2*IC(ncn(w_{i,m}, w_{j,n}))$$
$$= \log(P(ncn(w_{i,m}, w_{j,n}))/P(w_{i,m})) + \log(P(ncn(w_{i,m}, w_{j,n}))/P(w_{i,m}))$$
$$= \log P(ncn(w_{i,m}, w_{j,n}) \mid w_{i,m}) + \log P(ncn(w_{i,m}, w_{j,n}) \mid w_{i,m})$$

where *PW* is the path weight between $w_{i,m}$ and $w_{j,n}$, which means the more IC between concept links, the less the distance (more similar) in concept paths.

The simplified model can be explained as applying information content (Resnik 1995) on concept links, whereas the complicated, taking into account concept depth and density, can act as an improved version of Sussna's model. Their evaluation (Jiang and Conrath 1997) showed that the model made a marginal improvement to Resnik's model, with or without the depth and local density factors. Note that they trained and evaluated the model in the same dataset, hence over-tuned the optimal values for the depth and local density factors. Different settings of the depth and local density factors in their experiments made no significant difference in word similarity measures, which implied their negligible roles in calculating word similarity.

In a variant of this similarity model proposed by Lin (1997), $PW(w_{i,m}, w_{j,n})$ was defined in the form of $2*IC(ncn(w_{i,m}, w_{j,n}))/(IC(w_{i,m}) + IC(w_{j,n}))$, which functioned like Dice coefficient. Lin defined word similarity in the same form as Wu and Palmer (1994), and only substituted concept depths in their formula with concept ICs.

| Taxonomic similarity | Methods (abbr.) | Typical features | Semantic similarity: $Sim(w_{i,m}, w_{j,n})$ | Taxonomy | | | Statistics |
|---|---|---|---|---|---|---|---|
| | | | | Shortest path | Link type | Path type | Link type |
| Edge-counting | **EDGE**: Rada et al. (1989) | Simple edge-counting | $1/dist(w_{i,m}, w_{j,n})$ | ● | | | |
| | **LCH**: Leacock and Chodorow (1994, 1998) | | $-\log \dfrac{dist(w_{i,m}, w_{j,n})}{2*D}$ | ● | | | |
| | **WUP**: Wu and Palmer (1994) | Concept depth | $\dfrac{2*dep(ncn(w_{i,m}, w_{j,n}))}{dep(w_{i,m}) + dep(w_{j,n})}$ | ● | ● | | |
| | **AGI**: Agirre et al. (1994) | | $1 \Big/ \sum_{k=1}^{\lvert sd(w_{i,m},w_{j,n}) \rvert} 1/dep(w_k)$ | ● | ● | | |
| | **SUS**: Sussna (1993) | Concept depth Path type Concept density | $\dfrac{2 \times Max(dep(w_{i,m}), dep(w_{j,n}))}{dist(w_{i,m} \to_r w_{j,n}) + dist(w_{j,n} \to_{r^{-1}} w_{i,m})}$ | | | ● | ● |
| | **HSO**: Hirst and St-Onge (1997) | Path type | $C - dist(w_{i,m}, w_{j,n}) - K*dir(w_{i,m}, w_{j,n})$ | | | ● | |
| | **YP**: Yang and Powers (2005, 2006) | Hybrid utilization of concept specificity | $\alpha_t \times \beta_t^{sd(w_{i,m}, w_{j,n})-1}$ | ● | ● | ● | |
| Information content | **RES**: Resnik (1995) | Information content | $IC(ncn(w_{i,m}, w_{j,n})) = -\log P(ncn(w_{i,m}, w_{j,n}))$ | ● | | | ● |
| | **JCN**: Jiang and Conrath (1997) | Statistics-enhanced linkage strength | $2*IC(ncn(w_{i,m}, w_{j,n})) - (IC(w_{i,m}) + IC(w_{j,n}))$ | ● | | | ● |
| | **LIN**: Lin (1997) | | $2*IC(ncn(w_{i,m}, w_{j,n}))/(IC(w_{i,m}) + IC(w_{j,n}))$ | ● | | | ● |

Table 1: Summarization of the taxonomic similarity measures.

### 3.5. *Summary*

On the assumption of representing word meanings in the hand-crafted semantic networks, taxonomic similarity measures, as listed in Table 1, can work effectively in predicting semantic similarity, and can be generalized with three factors: the shortest path length, the link type, and the path type, among which the shortest path length makes a major contribution to yield semantic similarity. To calculate the shortest path length, EDGE and LCH only hypothesize the uniform distance with the simple edge-counting, whereas WUP, YP, and information content: RES, JCN, and LIN also leverage taxonomy structures and word usage frequencies to fine-tune the uniform distance. Besides, HSO and YP factor in the path type in mixing the IS-A and HAS-A hierarchies to enrich semantic connections. Aside from the shortest path length in semantic networks, the key difference among taxonomic similarity methods is how to estimate concept specificity, either using network structures in edge-counting or using concept frequencies in information content.

## 4. Neural network embeddings

Vector space models (Turney and Pantel 2010), generating distributional similarity, can calculate semantic similarity alternatively on the premise of similar words sharing similar contexts (Harris 1985). The growing research on distributional semantics has focused on unified (Bengio et al. 2003, Collobert et al. 2011, Mikolov et al. 2013b, Pennington et al. 2014) and contextualized (Devlin et al. 2018, Howard and Ruder 2018, Peters et al. 2018a, Radford et al. 2018) neural network embeddings, together with the hybrid method of retrofitting concept relationships into NNEs (Faruqui et al. 2015, Wieting et al. 2015, Mrkšić et al. 2016, Mrkšić et al. 2017, Ponti et al. 2018, Shi et al. 2019).

### 4.1. *Distributional semantics*

NNEs or the prediction-based embeddings—as a cornerstone of state-of-the-art distributional semantics—have significantly improved downstream NLP tasks (Collobert et al. 2011, Baroni et al. 2014). In contrast to the traditional counting-based distributional semantics such as Pointwise Mutual information (PMI) and Latent Semantic Analysis (LSA) (Deerwester et al. 1990, Turney 2001, Turney and Pantel 2010), NNEs can yield much denser word vectors with the reduce dimensionalities in an unsupervised way of learning. Note that the dimensionalities of NNEs can impose a varied impact on their quality, which are often customized for different needs. However, the optimal NNEs, training results of dimensionality selection on one similarity task, often generalized poorly when testing on other similarity tasks or applications (Yin and Shen 2018, Raunak et al. 2019). As for prediction-based embeddings, we aim to examine the publicly available NNEs that have been well trained and widely adopted for transfer learning in NLP. Since optimizing dimensionality for NNEs is not a primary goal in this research, we keep their original sizes of dimensions untouched. We mainly embraced three types of NNEs in the evaluation.

### 4.2. *Unified NNEs*

We selected two NNEs: Skip-gram with Negative Sampling (SGNS[1]) (Mikolov et al. 2013a, Mikolov et al. 2013b) and GloVe[2] (Pennington et al. 2014) for comparison, along with another well-known SGNS-based model—fastText[3] (Bojanowski et al. 2017). fastText aims to solve the widespread out-of-vocabulary problem in NNEs through sub-word or n-gram character embeddings. In the word similarity and analogy tasks, Levy et al. (2015) systematically contrasted SGNS and GloVe with two counting-based distributional semantics: PMI and truncated Singular Value Decomposition (SVD). They concluded that fine-tuning hyper-parameters during the NNEs' training might contribute to their superiority over the counting-based models. Note that word embeddings derived from these NNEs are unified, which implies that different senses of a word share a uniform representation without sense disambiguation. We repeatedly tested the NNEs using different sizes of dimensionalities in the evaluation and only reported their optimal results for brevity.

---

[1] Skip-gram (https://code.google.com/archive/p/word2vec/): pre-trained in the Word2Vec tool on Google News dataset (about 100B tokens) with 300 dimensions.
[2] GloVe (https://nlp.stanford.edu/projects/glove/): pre-trained on Wikipedia and Gigaword 5 (about 6B tokens) with 300 dimensions.
[3] fastText (https://fasttext.cc/docs/en/pretrained-vectors.html): pre-trained on the English part of multilingual Wikipedia, with 300 dimensions.

### 4.3. *Fusion of semantic relationships and NNEs*

NNEs inclined to elicit semantic relatedness rather than semantic similarity (Hill et al. 2015, Lê and Fokkens 2015) because they were trained unsupervisedly with word co-occurrences in a context window. They can hardly distinguish synonyms from antonyms, an easy task for taxonomic similarity models. Some studies further fed the advanced co-occurrence information such as syntactic dependency (Levy and Goldberg 2014a) and sentiment information among linguistic units (Tang et al. 2014) into training NNEs. However, the results showed that such enrichment processes only played a limited role in improving NNEs on similarity prediction.

Another way of improving NNEs is to incorporate semantic relationships in LKBs by pulling synonyms deliberately closer in a latent semantic space. For example, using synonymous pairs extracted from WordNet and the Paraphrase databases (Ganitkevitch et al. 2013), Yu and Dredze (2014) attempted to regularize the log-probability loss function during training NNEs; and Faruqui et al. (2015) designed an efficient way of retrofitting word embeddings after training NNEs. PARAGRAM (Wieting et al. 2015) only employed the Paraphrase databases in training a Skip-gram initialized word embedding.

Apart from post-processing or retrofitting NNEs with synonyms, Mrkšić et al. (2016) further took into account antonymous relationships—dubbed as counter-fitting (CF)—to push antonyms away during updating NNEs. They found that counter-fitting GloVe and PARAGRAM significantly improved their performances in predicting semantic similarity. Mrkšić et al. (2017) also employed multilingual semantic relationships in BabelNet (Navigli and Ponzetto 2012a) during counter-fitting NNEs and gained an improvement on the cross-lingual semantic similarity tasks. The procedure of CF (Mrkšić et al. 2016) is to first extract external semantic constraints: synonyms *S* and antonyms *A* from LKBs, which are then used to update the original vector space *R* through a contrastive loss *L*. To derive the retrofitted space $R^*$, CF defines *L* as follows:

$$L = L_S(B_S) + L_A(B_A) + L_P(R, R^*), \text{ where}$$

- $B_S$ and $B_A$ are two sets storing the external constraints *S* and *A*, respectively. For each synonymous or antonymous word-pair (*l*, *r*) in the sets, CF updates its corresponding vectors ($X'_l$, $X'_r$) in $R^*$ with backpropagation;

- $L_S(B_S) = \sum_{(l,r) \in B_S} \tau(\delta_{syn} - cos(X'_l, X'_r))$, calculating the loss $L_S$ to make synonymous vectors alike in $R^*$. $\tau(x) = max(0, x)$ is the hinge loss function. $\delta_{syn}$ denotes the maximum margin of similarity score that $X'_l$ and $X'_r$ can have. A part of the training objective of CF is to maximize distributional similarity between $X'_l$ and $X'_r$ to approach $\delta_{syn}$ or to pull $X'_l$ and $X'_r$ in $R^*$ close enough to one direction. $\delta_{syn}$ is adjustable for different tasks;

- $L_A(B_A) = \sum_{(l,r) \in B_A} \tau(\delta_{ant} + cos(X'_l, X'_r))$, calculating the loss $L_A$ to make antonymous vectors reciprocal in $R^*$. In contrast to $\delta_{syn}$, $\delta_{ant}$ is the minimum margin of similarity score that $X'_l$ and $X'_r$ can have. $\delta_{ant} = 0$ indicates that CF intends to train distributional similarity between $X'_l$ and $X'_r$ close to 0 or to push $l$ and $r$ in $R^*$ to a vertical direction;

- $L_P(R, R^*) = \sum_{i=1}^{|V|} \sum_{j \in Neg(i)} \tau \left( cos(X_i, X_j) - cos(X'_i, X'_j) \right)$, preserving distributional distance learned in $R$ when updating $R^*$ with semantic constraints of $B_S$ and $B_A$. $V$ is the vocabulary in $R$. For a word $i$ in $V$, CF first retrieves a group of its negative sample words ($Neg(i)$) that are within a threshold of distributional similarity between $i$ and $j$, and then keeps their respective semantic distance in $R$ and $R^*$ as same as possible.
- $L_S$, $L_A$, and $L_P$ can have different weights in retrofitting NNEs for a specific task, and other semantic relationships such as IS-A and PART-OF can also be incorporated in $L$.

More recent studies showed moderate success in using multiple LKBs to retrofit NNEs. For example, Speer and Chin (2016) utilized an ensemble method of merging GloVe and SGNS with the Paraphrase databases and ConceptNet (Speer and Havasi 2012); Ponti et al. (2018) employed generative adversarial networks (GANs) to combine WordNet and Roget's thesaurus with NNEs; Shi et al. (2019) demonstrated some positive effects of retrofitting paraphrase contexts into NNEs.

Given its significant gains in semantic similarity prediction, we chose PARAGRAM+CF[4] (Mrkšić et al. 2016)—imposing semantic constraints on PARAGRAM with antonyms in WordNet and the Paraphrase databases—to investigate the fusion effect of integrating semantic relationships into NNEs.

### 4.4. *Contextualized NNEs.*

In the unified (Bengio et al. 2003, Collobert et al. 2011, Mikolov et al. 2013b, Pennington et al. 2014) and hybrid (Faruqui et al. 2015, Wieting et al. 2015, Mrkšić et al. 2016) NNEs, a polysemous word shares a uniform embedding without any sense distinction. An ongoing research trend on NNEs has attended to cultivating contextualized embeddings using NLMs, e.g. CoVe (McCann et al. 2017), ELMo (Peters et al. 2018a), ULMFiT (Howard and Ruder 2018), GPT-2 (Radford et al. 2018), and BERT (Devlin et al. 2018). The contextualized embeddings leverage word order in contexts—often neglected in the static or unified NNEs—through deeper neural networks such as bi-directional RNN and Transformer with attention mechanism (Vaswani et al. 2017). They have greatly improved many natural language understanding tasks. For example, BERT—using both masked language modelling and sentence prediction in pre-training—had attained cutting-edge performance on some benchmark tasks such as GLUE (Wang et al. 2018) and SQuAD (Rajpurkar et al. 2016).

The impressive results of these contextualized NNEs (hereafter CNNEs in short) can be attributed to the utilization of deep and multi-layered neural architectures such as CNNs, RNNs, and Transformers. Such architectures can not only produce context-dependent word embeddings but also display various linguistic features in their respective layers (Peters et al. 2018b, Goldberg 2019, Tenney et al. 2019). Analogously, deep learning architectures on computer vision and ImageNet (Russakovsky et al. 2014) can learn multi-grained hierarchical features, from specific edges to general and abstractive patterns in images (Krizhevsky et al. 2012, Yosinski et al. 2014). For the linguistic features learned from BERT, Jawahar et al. (2019) demonstrated in a series of probing tasks that its lower layers could contain morphological information and the middle layers and upper layers could capture syntactic and semantic information, respectively. Ethayarajh (2019) also showed that the upper layers in ELMo, BERT, and GPT-2 might contain more task or context-specific information than the lower ones.

---

[4] PARAGRAM+CF (https://github.com/nmrksic/counter-fitting): initialized using Skip-gram trained on Wikipedia (about 1.8B tokens) with 300 dimensions.

Among these deep learning CNNEs, we chose BERT-base[5] rather than BERT-large in the evaluation. Even when BERT-base was pre-trained with a much less complicated structure than BERT-large (a sophisticated variant of BERT), it maintained competitive performance on some benchmark tasks. Throughout the evaluation, we used BERT to refer to BERT-base.

Unlike the unified word representations in NNEs, we used BERT to generate the sense representations of a polysemous word although such generation can only signify context-dependent word meanings at most. Directly feeding a word without its contexts into BERT would only retrieve implausible sense representations, equivalent to the unified NNEs in essence. Ideally, a synset's example sentences in WordNet can be a proxy to its contexts, but nearly 50% of the nouns and 7% of the verbs in the evaluation datasets lack of example sentences. We, therefore, replaced them with a synset's gloss as the input of BERT. For each synset of a word, we averaged the vectors of all the tokens in its corresponding gloss to yield an aggregated embedding, which can function as the sense embedding of a word. We then defined semantic similarity between a pair of words as the maximum of distributional similarity scores (cosine) across their sense embeddings in BERT, derived in the same way as the maximum of taxonomic similarity across different senses in Section 3. Given that the top layers in BERT are more context-specific and semantics-oriented, we first individually calculated the sense embeddings of a synset from the $9^{th}$ to $12^{th}$ layer, together with an average embedding on the four layers. Among these five embeddings, we only highlighted the best, achieved on the $11^{th}$ layer, in the evaluation.

## 5. Evaluating taxonomic and neural embedding models

Fairly comparing taxonomic and neural embeddings models in a single framework is not a relaxing task. In analysing the Brown corpus and British National Corpus (BNC), Kilgarriff (2004) concluded that both word and sense frequency held similar Zipfian (Zipf 1965) or Power-law distribution with an approximately constant product of frequency and rank. Likewise, in the evaluations of taxonomic similarity models through lexical semantic applications such as malapropism detection and correction (Budanitsky and Hirst 2006), WSD (Pedersen et al. 2005), and the predominant (first) sense prediction (McCarthy et al. 2004a, McCarthy et al. 2004b), it was shown that the incorporation of sense frequencies in calculating information content could introduce a bias against the edge-counting models that only employ infrastructure features in semantic networks. As the Zipfian distribution of word senses may be in favour of the predominant or first sense of a word, information content may estimate concept specificity inaccurately. By contrast, edge-counting may show no preferences for sense ranks in yielding similarity from all the existing paths.

Since it is hardly to avoid the interference of skewed sense distribution or word frequency on evaluating taxonomic and distributional similarity models in the lexical semantic applications, we assess them by contrasting their similarity judgements with human-conceived scores. Mohammad and Hirst (2012) also recommended a direct comparison of distributional similarity with human similarity ratings, which can indicate its distinctiveness in contrast with taxonomic similarity measures. We compare the following models in the evaluation, which include:
- taxonomic similarity (as listed in Table 1): edge-counting with EDGE (Rada et al. 1989), LCH (Leacock and Chodorow 1994, 1998), WUP (Wu and Palmer 1994), HSO (Hirst and St-Onge 1997), and YP (Yang and Powers 2005, 2006), together with information content with RES (Resnik 1995), JCN (Jiang and Conrath 1997), and LIN (Lin 1997).

---
[5] BERT-base (https://github.com/google-research/bert): its uncased variant features 12 layers, 768 hidden units, 12 attention heads, and 110M parameters.

- distributional similarity: unified NNEs with SGNS (Mikolov et al. 2013a, Mikolov et al. 2013b), GloVe (Pennington et al. 2014), and fastText (Bojanowski et al. 2017), hybrid NNEs with PARAGRAM+CF (Mrkšić et al. 2016), together with contextualized NNEs with BERT (Devlin et al. 2018).

### 5.1. *Human similarity judgements*

Human similarity judgements are often acquired in a context-free setting where it is common for the lack of corresponding contextual hints to help disambiguate word meanings. With both context and non-context aiding, Resnik and Diab (2000) designed a pilot dataset only consisting of 48 verb pairs. In their pioneering work, they found that taxonomic similarity models (i.e. EDGE, RES, and LIN) worked likewise in the non-context setting, whereas they all improved to some degree in the context setting. Their results indicated that human subjects might comprehend word meanings more arbitrarily when corresponding contextual hints were absent. Given the complexity and laboriousness of constructing datasets in the context setting, nearly all the current datasets in use, e.g. SimLex-999 (Hill et al. 2015) and SimVerb-3500 (Gerz et al. 2016), are context-free. They were dedicatedly designed for measuring semantic similarity rather than semantic association (Hill et al. 2015). Note that hereafter to name a dataset, the initial numbers stand for its size, followed by its PoS tag. We only collected some context-free datasets for evaluation. They consisted of two datasets on nouns (N): 65_N (Rubenstein and Goodenough 1965) and the similarity subset of 201_N (Agirre et al. 2009)—extracted from 353_N (Finkelstein et al. 2001); two on verbs (V): 130_V (Yang and Powers 2006) and 3000_V—the test part of SimVerb-3500; only one containing both nouns, verbs, and adjectives (ALL)—999_ALL, i.e. SimLex-999. We further extracted two subsets from 999_ALL: 666_N and 222_V.

### 5.2. *Definition of an upper bound*

Past studies defined an upper bound in evaluation via *average correlation coefficients* (Spearman's $\rho$ or Pearson's $r$) among pairwise annotators or between a leave-one-out annotator and group average. Such a definition on the upper bound is often problematic because it can be surpassed or reached with a proximity to humans. For example, using their hybrid taxonomic similarity model, Yang and Powers (2005) achieved $r = 0.92$ on 30_N (Miller and Charles 1991), surpassing the upper bound of $r = 0.90$ in the pairwise agreement. Recski et al. (2016) designed a regression model combining multiple distributional similarity models, taxonomy features, and features extracted from a concept dictionary. They achieved $\rho = 0.76$ on 999_ALL, well exceeding the upper bound of $\rho = 0.67$ in the pairwise agreement and close to the group average of $\rho = 0.78$. These excellent results need to be interpreted with due care because it does not mean their methods (Yang and Powers 2005, Recski et al. 2016) have gone beyond human similarity judgements. It only suggests that they could achieve more accurate results than most of the individual annotators on the specific datasets. As noticed by Lipton and Steinhardt (2018), before we can understand the internal mechanism of human similarity judgements, we should employ extreme caution in using "human-level" terms in describing computational models or claiming their supremacy over humans. Therefore, an upper bound needs to be established properly.

We suggest that the upper bound should be taken as an average agreement among different human groups on the same task (groupwise agreement), through which we can correctly set up a maximum on similarity judgements. For example, Rubenstein and Goodenough (1965) hired 51 college undergraduates to evaluate their well-known 65_N; and

Miller and Charles (1991) extracted 30 pairs from 65_N, and repeated a similar experiment with 38 subjects. As for 65_N and 30_N, where human similarity scores are evenly distributed from high to low, the correlation coefficient between the mean human ratings of the two groups is $r = 0.97$ or $\rho = 0.95$. Moreover, Resnik (1995) replicated an experiment on the 28 pairs extracted from 30_N. He hired 10 subjects and achieved $r = 0.96$ or $\rho = 0.93$ between the mean human ratings of 28_N and 30_N. Therefore, the upper bound for judging noun similarity should approximate $r = [0.96, 0.97]$ or $\rho = [0.93, 0.95]$. Analogously, to estimate the upper bound on the verb task, we calculated a groupwise agreement on the 170 verb pairs that occurred in both 999_ALL (50 participants) and SimVerb-3500 (843 participants), which stands at $r = 0.92$ or $\rho = 0.91$ (Gerz et al. 2016). Given the reasonably large number of participants in the experiments, these values can serve as the reliable upper bounds in the evaluation.

### 5.3. *Results on intrinsic evaluation*

Note that in the following sections, we solely reported Spearman's rank correlation on head-to-head comparisons as it makes no assumptions of linear dependence on similarity scores.

As indicated in the heat map of Figure 2, the group performance (mean 0.56, 95% confidence interval (CI) 0.47 to 0.65) of the taxonomic similarity measures was on average better than that (mean 0.47, 95% CI 0.34 to 0.62) of SGNS, fastText, GloVe, and BERT. This finding is in line with the previous studies on NNEs (Hill et al. 2015, Lê and Fokkens 2015). Except for PARAGRAM+CF, the two groups of similarity measures could be roughly divided into two corresponding clusters in the dendrogram of Figure 2, apparently showing distinctive patterns over the seven datasets. PARAGRAM+CF with a mean correlation of 0.71 (95% CI 0.68 to 0.75) was substantially different from both groups in Figure 2.

Moreover, edge-counting and information content showed no significant pattern differences in Figure 2 (Friedman test: $\chi^2 = 7.65$, $p = 0.37$). The result partly supports our hypothesis on the prime role of the shortest path length in calculating taxonomic similarity and suggests that only a minor role of the link or path type can play in adjusting the uniform distance. Except for PARAGRAM+CF, no significant differences existed among NNEs (Friedman test: $\chi^2 = 3.51$, $p = 0.32$).

PARAGRAM+CF tends to have a more consistent similarity prediction with a standard deviation of 0.06 (95% CI 0.02 to 0.08) than other measures across the seven datasets, and performed competently well above others on 65_N, 666_N, 222_V, and 3000_V. It also outperformed others on 999_ALL, the only dataset containing both nouns, verbs, and adjectives. In general, PARAGRAM+CF, retrofitting multiple semantic relationships in LKBs into NNEs, improved the capability of distributional vector models in yielding semantic similarity, which corroborates previous findings in the literature (Faruqui et al. 2015, Mrkšić et al. 2016, Mrkšić et al. 2017).

Concerning a method's divergence on the group average performance, we first defined its mean correlation ratio (MCR in short) from noun to verb task to quantify performance variability. As shown in the column chart in Figure 2, the MCR of PARAGRAM+CF (1.10) was only larger than that of JCN (1.00) and LIN (1.01), which possibly indicates a similar level of variability in similarity prediction. EDGE, WUP, LCH, RES, HSO, and YP ([1.24, 1.48]) had relatively lower MCRs than SGNS, GloVe, fastText, and BERT ([1.47, 1.96]), which implies that taxonomic similarity could be on average more stable than NNEs. Overall, the MCRs also indicate that nearly all similarity measures (except for JCN and LIN) deteriorated by a certain degree from noun to verb task. The overall performance of these methods on the noun datasets (mean 0.63, 95% CI 0.60 to 0.66; standard deviation 0.06, 95% CI 0.03 to 0.08) was better than it on the verb ones (mean 0.48, 95% CI 0.43 to 0.53; standard deviation 0.10,

95% CI 0.06 to 0.13). The result suggests that it may be more challenging to compute verb similarity than noun similarity, to which a hybrid use of neural embeddings and semantic relations, e.g. PARAGRAM+CF, may provide a solution.

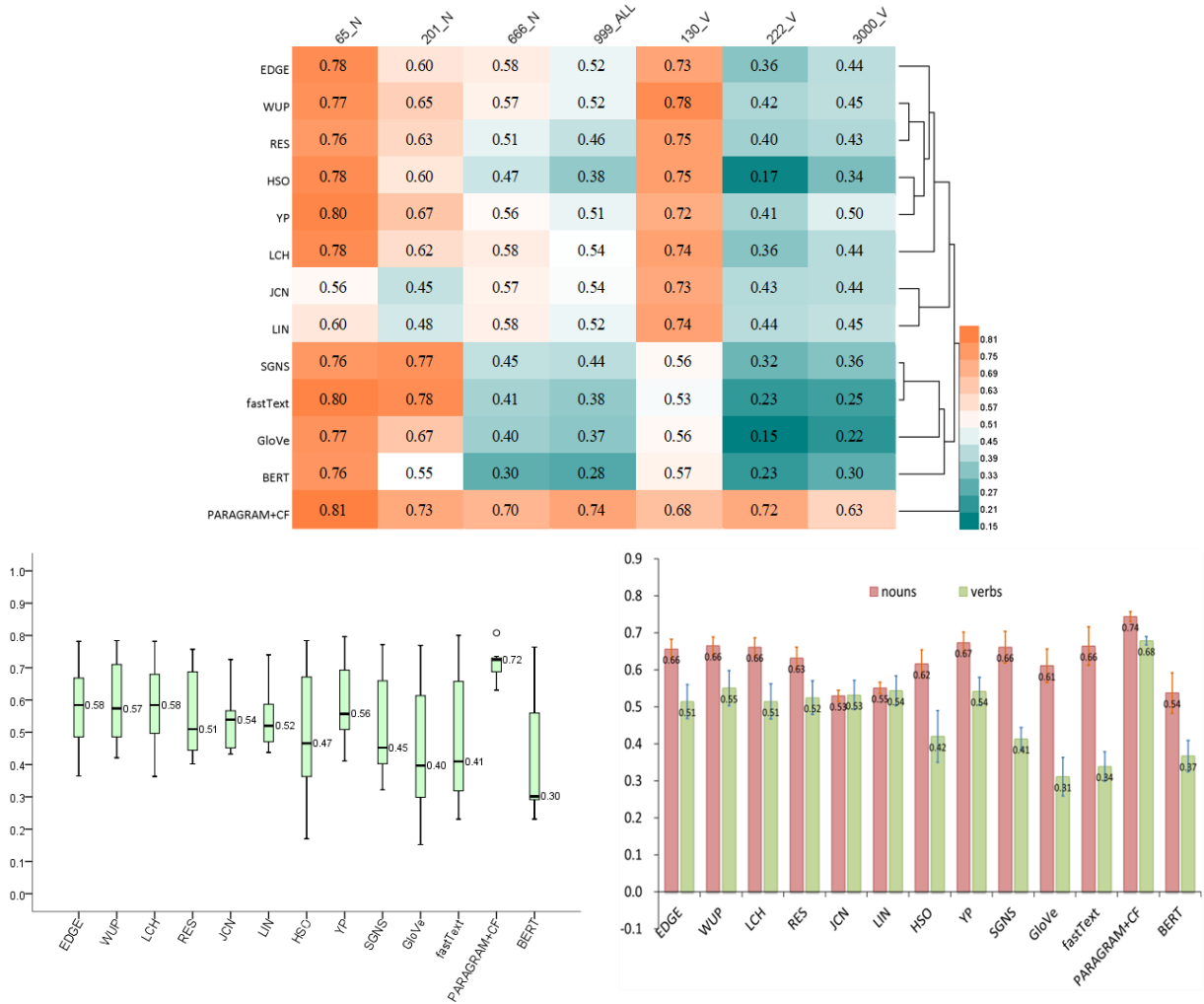

Figure 2: Comparison of the taxonomic and distributional similarity measures on the seven benchmark datasets (in terms of Spearman's ρ). We attached the heat map (top left) with a dendrogram of the measures. We applied a hierarchical clustering using average linkage and Pearson distance on the measures. The boxplot (bottom left) shows the variation of each measure's correlations across the seven datasets. The column chart (bottom right) further indicates the performance variation of each measure on the group-aggregated level of nouns (65_N, 201_N, and 666_N) and verbs (130_V, 222_V, and 3000_V).

### 5.4. *Noun and verb similarity*

Apart from a performance overview of these measures on the seven benchmark datasets, we further pinpoint their detailed differences on yielding semantic similarity on nouns and verbs in the datasets: 666_N and 3000_V, respectively. We propose three criteria: word frequency, polysemy degree, and similarity intensity to investigate their effects on similarity prediction.

5.4.1 *Word frequency*

Word frequency is highly associated with synset ranking in WordNet. Frequency statistics also have a crucial application in similarity calculation, such as factoring in concept specificity with information content and pre-training neural embeddings. To investigate word frequency's effect on similarity judgements, we first set up three intervals that cover word usages from low, medium to high-frequency ranges, sampled from BNC. We then extracted word pairs from each dataset with their word occurrence numbers that are both sat in one of the three intervals.

| | Nouns | | | | Verbs | | | |
|---|---|---|---|---|---|---|---|---|
| | (0-3,000): 61 | (3,000-10,000): 104 | 10,000+: 70 | 666_N | (0-1,000): 449 | (1,000-5,000): 355 | 5,000+: 275 | 3000_V |
| EDGE | 0.61 | 0.57 | 0.57 | 0.58 | 0.49 | 0.44 | 0.48 | 0.44 |
| WUP | 0.59 | 0.53 | 0.56 | 0.57 | 0.41 | 0.48 | 0.49 | 0.45 |
| LCH | 0.62 | 0.57 | 0.57 | 0.58 | 0.47 | 0.45 | 0.48 | 0.44 |
| RES | 0.41 | 0.39 | 0.38 | 0.51 | 0.35 | 0.45 | 0.47 | 0.43 |
| JCN | 0.50 | 0.55 | 0.55 | 0.57 | 0.42 | 0.46 | 0.43 | 0.44 |
| LIN | 0.50 | 0.53 | 0.54 | 0.58 | 0.40 | 0.46 | 0.49 | 0.45 |
| HSO | 0.53 | 0.36 | 0.37 | 0.47 | 0.38 | 0.36 | 0.37 | 0.34 |
| YP | 0.60 | 0.61 | 0.60 | 0.56 | 0.50 | 0.47 | 0.61 | 0.50 |
| SGNS | 0.42 | 0.32 | 0.33 | 0.45 | 0.37 | 0.36 | 0.40 | 0.36 |
| GloVe | 0.42 | 0.32 | 0.25 | 0.40 | 0.24 | 0.25 | 0.29 | 0.22 |
| fastText | 0.34 | 0.37 | 0.24 | 0.41 | 0.32 | 0.23 | 0.29 | 0.25 |
| PARAGRAM+CF | 0.56 | 0.66 | 0.81 | 0.70 | 0.59 | 0.66 | 0.68 | 0.63 |
| BERT | 0.29 | 0.28 | 0.32 | 0.30 | 0.32 | 0.29 | 0.30 | 0.30 |

Table 2: Word frequency's effects on yielding semantic similarity. For the noun pairs in 666_N (left), we first divided their frequency range into three intervals: less than 3,000; between 3,000 and 10,000; and more than 10,000. We then individually acquired 61, 104, and 70 noun pairs that can satisfy the frequency requirements of the three intervals. Likewise, in 3000_V (right), 449, 355, and 275 pairs of verbs were respectively extracted for the three frequency intervals: less than 1,000; between 1,000 and 5,000; and more than 5,000. We also added the overall results on 666_N and 3000_V for reference.

As shown in Table 2, EDGE and LCH—only employing the shortest path length in WordNet—were little sensitive to the frequency variations on both datasets (SE $\leq$ 0.01), and each of them was not significantly different for the frequency ranges on 666_N (one-way ANOVA: EDGE: $F(2, 232) = 1.06$, $P = 0.35$; LCH: $F(2, 232) = 1.74$, $P = 0.18$ ) and 3000_V (one-way ANOVA: EDGE: $F(2, 1072) = 0.15$, $P = 0.86$; LCH: $F(2, 1072) = 0.32$, $P = 0.72$). Although the link type and path type have impacts on adjusting the uniform distance, information content: RES, JCN, and LIN, together with edge-counting: WUP, HSO, and YP, showed no significant performance variations over the frequency intervals, as determined by one-way ANOVA ($P > 0.05$). Except for PARAGRAM+CF, YP also attained the best in each of the intervals with a consistent performance on 666_N (SE = 0.00), but it showed a little fluctuation on 3000_V (SE = 0.03) despite no significance (one-way ANOVA: $F(2, 1072) = 0.34$, $P = 0.72$).

NNEs including SGNS, GloVe, and fastText, along with PARAGRAM+CF, were more sensitive to the frequency variations on 666_N (SE: [0.03, 0.06]) than on 3000_V (SE: [0.01, 0.02]), but such sensitivity was not significant, according to one-way ANOVA ($P > 0.05$). However, fastText (one-way ANOVA: $F(2, 1072) = 8.27$, $P = 0.00$), PARAGRAM+CF (one-way ANOVA: $F(2, 1072) = 5.60$, $P = 0.00$), and GloVe (one-way ANOVA: $F(2, 1072) = 4.61$, $P = 0.01$) showed significant performance variations on 3000_V, except for SGNS (one-way ANOVA: $F(2, 1072) = 1.90$, $P = 0.15$). Overall, PARAGRAM+CF achieved better results than YP, the best in edge-counting, excluding the low frequency range on 666_N. In contrast to other NNEs, BERT was less subject to the frequency change (SE = 0.01, one-way ANOVA: $F(2, 232) = 2.09$, $P = 0.13$ on 600_N and SE = 0.01, $F(2, 1072) = 0.38$, $P = 0.69$ on 3000_V).

Note that apart from the five datasets in our evaluation, mainly consisting of frequent words in generic domains, there are other benchmark similarity datasets such as RW-2034 (Luong et al. 2013) and CARD-660 (Pilehvar et al. 2018). They were particularly designed to cover infrequent or rare words, which were retrieved from extremely larger sizes of corpora than BNC. They are often taken to validate NNEs' quality, especially in dealing with out-of-vocabulary (OOV) words. Although fastText and BERT can address the OOV issue through sub-word encoding, the other NNEs in the evaluation, along with taxonomic measures, are still subject to the limited sizes of their vocabularies. We therefore extracted two subsets of 349 and 52 word pairs from RW-2034 and CARD-660, respectively, in which all the measures can yield corresponding similarity scores on any target pair.

As shown in Figure 3, taxonomic similarity achieved better results on CARD-660 (mean±SD: 0.46±0.11) than on RW-2034 (0.27±0.05), whereas NNEs performed adversely with 0.01±0.07 on CARD-660 and 0.45±0.16 on RW-2034. PARAGRAM+CF (0.58), along with SGNS (0.58), surpassed the other measures on RW-2034, but it plummeted to 0.15 on CARD-660, sharply lower than taxonomic measures and modestly higher than the other NNEs. In line with the results on the low frequency ranges on 666_N and 3000_V in Table 2, taxonomic similarity measures were relatively stable on calculating similarity on rare words, but NNEs fluctuated sharply. Given that our findings are based on a limited number of rare words, and the upper bounds (around 0.43 and 0.90, respectively) on RW-2034 and CARD-660 are substantially different, the results on a small scale of coverage should therefore be treated with considerable caution. For reasons of space, a further investigation into rare words are reserved for future work.

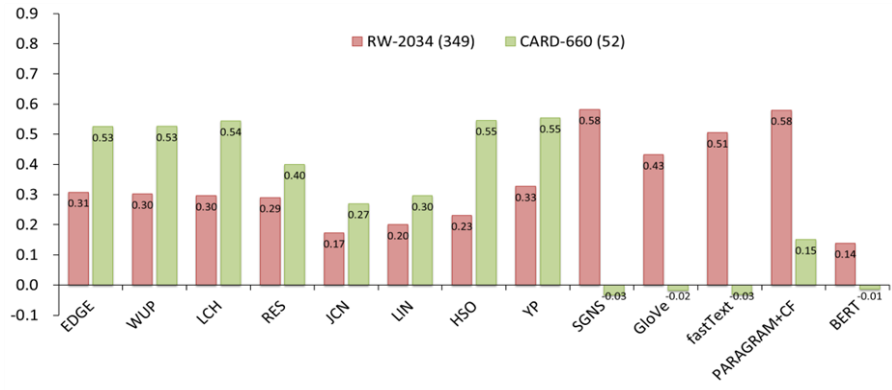

Figure 3: Yielding semantic similarity on rare words.

Our findings suggest that taxonomic similarity may be neutral to the variations of word frequency, partly because edge-counting mainly relies on semantic networks to calculate the shortest path length; partly because information content may ineffectively estimate concept specificity in adjusting the uniform distance. As the neural embeddings are the results of pre-training neural networks with co-occurrences, their performances may be susceptible to the frequency variations. Especially for words with higher frequencies, their corresponding neural embeddings may put them into a better position in yielding semantic similarity.

5.4.2 *Polysemy degree*

In WordNet[6], nouns have 2.79 senses on average, and verbs 3.57. With the increase of a word's polysemy degree, sense disambiguation would become more demanding, and searching the shortest path would engage in more path options in a taxonomy. It may result in a potential misestimation of taxonomic similarity and incorrect identification of authentic word senses in contexts. Additionally, the more senses a word owns implies the more idiosyncratic context patterns it will have. Assembling these patterns in a uniform representation may undermine the representability of distributional vector models in deriving semantic similarity.

---

[6] https://wordnet.princeton.edu/documentation/wnstats7wn

As shown in Table 3, all the taxonomic similarity measures worsened off while working on the highly polysemous nouns. However, one-way ANOVA indicates that their performance variations (SE = 0.03 on average) on the polysemy degree were not significant ($P > 0.05$). By contrast, they were less varied on the polysemous verbs (SE = 0.01 on average) but with a significant difference ($P \ll 0.05$, one-way ANOVA). The result implies that the highly polysemous nouns tend to have more complex interconnections in semantic networks, which can be problematic to search the shortest path effectively and then yield similarity scores correctly; the shallow and simple verb hierarchy may lessen the impact of polysemy degree on similarity calculation.

Interestingly, the polysemy degree tallied with the performance of each NNE on 3000_V, that is, the less sense number or weaker polysemy degree indicates the more accurate outcomes on similarity prediction ($P \ll 0.05$, one-way ANOVA). This finding is in line with the previous work on the polysemy effects on distributional similarity (Gerz et al. 2016). Among NNEs, PARAGRAM+CF performed the best on each level of polysemy degree. The sense or contextualized embeddings, generated by BERT, outperformed the unified NNEs: SGNS, GloVe, and fastText on the medium and high ranges of polysemy degree on 3000_V, but not on the high range on 666_N.

|  | Nouns | | | | Verbs | | | |
| --- | --- | --- | --- | --- | --- | --- | --- | --- |
|  | (0-2): 91 | (2-4): 152 | 4+: 102 | 666_N | (0-3): 344 | (3-8): 480 | 8+: 319 | 3000_V |
| EDGE | 0.60 | 0.68 | 0.42 | 0.58 | 0.56 | 0.45 | 0.55 | 0.44 |
| WUP | 0.60 | 0.67 | 0.39 | 0.57 | 0.53 | 0.48 | 0.47 | 0.45 |
| LCH | 0.60 | 0.69 | 0.43 | 0.58 | 0.55 | 0.44 | 0.53 | 0.44 |
| RES | 0.54 | 0.52 | 0.28 | 0.51 | 0.52 | 0.47 | 0.46 | 0.43 |
| JCN | 0.54 | 0.58 | 0.42 | 0.57 | 0.49 | 0.45 | 0.50 | 0.44 |
| LIN | 0.55 | 0.59 | 0.40 | 0.58 | 0.53 | 0.49 | 0.51 | 0.45 |
| HSO | 0.50 | 0.61 | 0.33 | 0.47 | 0.48 | 0.39 | 0.31 | 0.34 |
| YP | 0.54 | 0.68 | 0.50 | 0.56 | 0.60 | 0.54 | 0.55 | 0.50 |
| SGNS | 0.25 | 0.51 | 0.40 | 0.45 | 0.54 | 0.35 | 0.15 | 0.36 |
| GloVe | 0.16 | 0.49 | 0.40 | 0.40 | 0.46 | 0.20 | -0.02 | 0.22 |
| fastText | 0.19 | 0.42 | 0.39 | 0.41 | 0.38 | 0.16 | 0.08 | 0.25 |
| PARAGRAM+CF | 0.59 | 0.74 | 0.69 | 0.70 | 0.69 | 0.61 | 0.59 | 0.63 |
| BERT | 0.29 | 0.52 | 0.20 | 0.30 | 0.45 | 0.40 | 0.33 | 0.30 |

Table 3: Polysemy degree effects on judging semantic similarity. On 666_N (left), we divided polysemy degree into three intervals: less than or equal to 2 in sense numbers, between 2 and 4, and more than 4, in which we finally retrieved 91, 152, and 102 noun pairs, respectively. On 3000_V (right), we retrieved 344, 480, and 319 verb pairs according to the sense number intervals: less than 3, between 3 and 8, and more than 8, respectively. We also added the overall results on 666_N and 3000_V for reference.

Overall, in Table 3, when words become less polysemous or conceptually coarse-grained, both taxonomic similarity and neural embeddings can perform better on similarity prediction. Retrofitting NNEs with hand-crafted semantic networks can enhance distributional vector models in predicting semantic similarity on different levels of polysemy degree.

### 5.4.3 *Similarity intensity*

The ten-point Likert scale similarity scores in the gold-standard datasets, ranging from the highest (10) for synonyms or near-synonyms to the lowest for antonyms (0), can signify not only words' interchangeability in contexts but also their semantic nuances. In leveraging hand-crafted semantic relationships, taxonomic similarity can set up different weighting factors such as the link and path type to approach similarity intensity appropriately. By contrast, under the premise of similar words sharing similar contexts (Firth 1957), distributional similarity only depends on co-occurrence patterns in estimating similarity intensity. As synonyms and antonyms often share similar contextual patterns, distributional similarity may not differentiate them effectively without the aid of additional resources and attempts to score them both highly related.

As for 666_N in Table 4, all the measures worked significantly better in the low range of similarity intensity (mean±SE: 0.53±0.01) than in the medium (0.15±0.02) and high (0.29±0.02) ranges ($P \ll 0.05$, one-way ANOVA). Moreover, in the medium range, all the taxonomic similarity measures achieved higher correlation values than the neural embeddings, except for PARAGRAM+CF. Results on 3000_V in Table 4 also show that PARAGRAM+CF outperformed all other measures across the three ranges; on the high range, except for PARAGRAM+CF, all the measures achieved a significant net gain ($P \ll 0.05$, one-way ANOVA). Overall, all the measures, except for HSO, performed worst in the medium range on 666_N and 3000_V.

Our findings indicate that the taxonomic similarity measures failed to perform consistently on different levels of similarity intensity, which may be partly caused by the synset organization in WordNet. For example, on the high range of similarity intensity, only 15.6%, 26.6%, and 25.6% of word pairs (a total of 67.8%) in 666_N have the shortest path length of zero (synonyms), one, and two links in edge-counting, respectively; and so do 27.9%, 30.4%, and 18.5% of word pairs (a total of 76.9%) in 3000_V. The domain and coverage of existing LKBs, purpose-built by lexicographers, only recorded a relatively stable vocabulary knowledge (Kilgarriff 1997, Kilgarriff and Yallop 2000), which are diversified on the compilation of synonyms or near-synonyms. WordNet only organized paradigmatic relations into a fine-grained semantic network where the words in a synset node strictly conform to the notion of semantic similarity or interchangeability in contexts (Kilgarriff and Yallop 2000). Conversely, as for Roget's tree structure in both paradigmatic and syntagmatic relations, each node contains highly related words. The unbalanced distribution of highly similar words in WordNet may impose a constraint on the effect of edge-counting, dampening the conditioning effects of the link and path type on the uniform distance. The hybrid use of concept relationships and distributional semantics, e.g. PARAGRAM+CF, may address the unbalanced and outdated structure issue of LKBs in deriving semantic similarity.

|  | Nouns | | | | Verbs | | | |
|---|---|---|---|---|---|---|---|---|
|  | (0-3): 200 | (3-6): 267 | (6-10): 199 | 666_N | (0-3): 1,089 | (3-6): 1,033 | (6-10): 878 | 3000_V |
| EDGE | 0.49 | 0.26 | 0.34 | 0.58 | 0.10 | 0.10 | 0.31 | 0.44 |
| WUP | 0.55 | 0.23 | 0.32 | 0.57 | 0.14 | 0.07 | 0.29 | 0.45 |
| LCH | 0.51 | 0.26 | 0.34 | 0.58 | 0.11 | 0.08 | 0.30 | 0.44 |
| RES | 0.52 | 0.14 | 0.25 | 0.51 | 0.15 | 0.06 | 0.24 | 0.43 |
| JCN | 0.53 | 0.15 | 0.35 | 0.57 | 0.13 | 0.10 | 0.28 | 0.44 |
| LIN | 0.57 | 0.13 | 0.34 | 0.58 | 0.14 | 0.11 | 0.30 | 0.45 |
| HSO | 0.55 | 0.12 | 0.21 | 0.47 | 0.03 | 0.09 | 0.25 | 0.34 |
| YP | 0.49 | 0.18 | 0.33 | 0.56 | 0.17 | 0.12 | 0.32 | 0.50 |
| SGNS | 0.56 | 0.06 | 0.22 | 0.45 | 0.19 | 0.07 | 0.25 | 0.36 |
| GloVe | 0.59 | 0.06 | 0.21 | 0.40 | 0.14 | -0.03 | 0.20 | 0.22 |
| fastText | 0.59 | 0.03 | 0.26 | 0.41 | 0.14 | 0.05 | 0.18 | 0.25 |
| PARAGRAM+CF | 0.51 | 0.28 | 0.46 | 0.70 | 0.45 | 0.19 | 0.36 | 0.63 |
| BERT | 0.45 | 0.05 | 0.16 | 0.30 | 0.12 | 0.01 | 0.21 | 0.30 |

Table 4: Similarity intensity effects on judging semantic similarity. On 666_N (left), we divided the gold-standard similarity scores into three intensity ranges: less than 3, between 3 and 6, and more than 6, and each range contains 200, 267, and 199 pairs, respectively. On 3000_V (right), 1,089, 1,033, and 878 pairs were individually extracted from the three ranges. As for each method, the dark red error bar indicates its mean and standard error. We also added the overall results on 666_N and 3000_V for reference.

### 5.4.4 *Results aggregation*

After investigating these measures under the conditions of word frequency, polysemy degree, and similarity intensity on 666_N and 3000_V, we assembled their results to further compare their patterns in similarity prediction. Figure 4 displays that the group: EDGE, WUP, LCH, JCN, and LIN, forming a hierarchical cluster, shared a similar pattern of performances on 666_N and 3000_V (Friedman test: $\chi^2 = 7.50$, $p = 0.11$), so did the group of RES and HSO (Friedman test: $\chi^2 = 1.03$, $p = 0.60$). As for information content, no significant pattern differences existed between JCN and LIN (Friedman test: $\chi^2 = 1.00$, $p = 0.45$). The neural embedding group, including SGNS, GloVe, fastText, and BERT, also demonstrated a significantly similar pattern (Friedman test: $\chi^2 = 0.91$, $p = 0.63$), but PARAGRAM+CF distanced itself from any group, which is in good agreements with our findings in Figure 2.

The aggregated results in Figure 4 are consistent with the previous findings, which in general suggest that the taxonomic similarity measures can outperform the unified and contextualized neural embeddings in similarity prediction. Additionally, the simple edge-counting can effectively yield semantic similarity without using the conditioning factors on the uniform distance. PARAGRAM+CF, retrofitting NNEs with multiple semantic relations such as synonyms and antonyms, can pave a new way of combining taxonomic and distributional similarity to improve semantic similarity prediction.

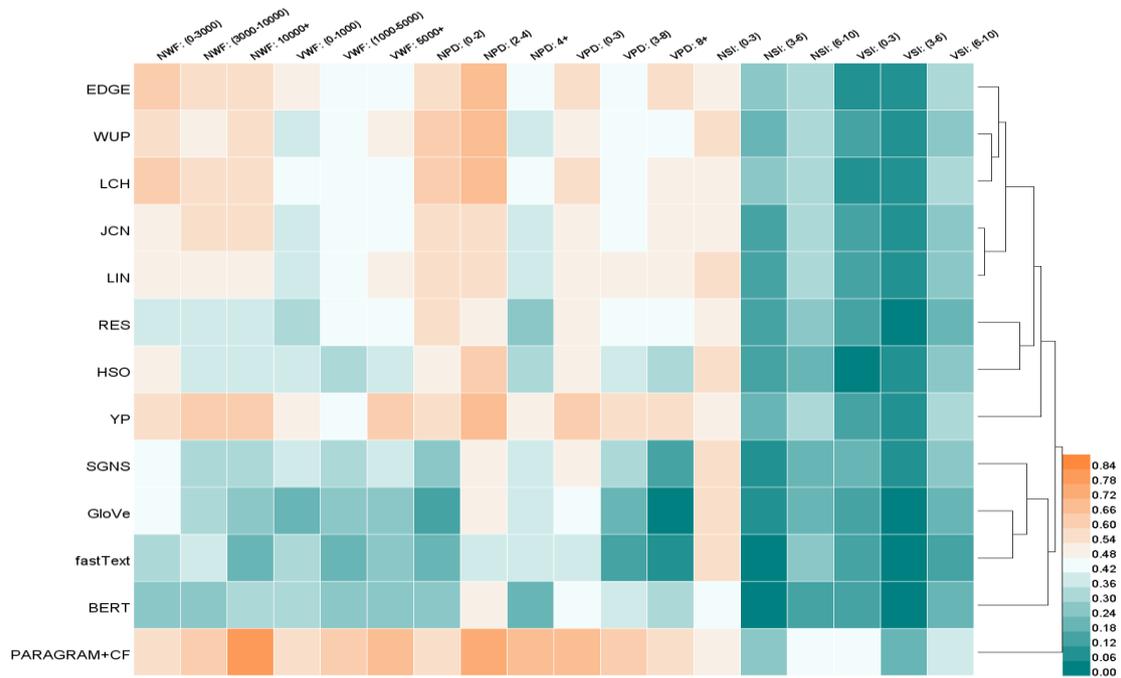

Figure 4: An overview of the effects of word frequency (WF), polysemy degree (PD), and similarity intensity (SI) on similarity judgements. For each column name, N and V denote the two datasets: 666_N and 3000_V, respectively. We added a dendrogram of the similarity measures, using average linkage and Euclidean distance.

## 6. Resemblance of similarity prediction patterns

Apart from examining these measures with the gold-standard human scores on similarity judgements, we further compared them through calculating their mutual correlations on 666_N and 3000_V to disclose their inter-proximity on prediction patterns.

### 6.1. *Taxonomic similarity*

As shown in Figure 5, all the taxonomic similarity measures worked likewise on 666_N (mean 0.80, 95% CI 0.76 to 0.83) and 3000_V (mean 0.79, 95% CI 0.76 to 0.82), among which EDGE,

WUP, and LCH were extraordinarily close with a mean $\rho$ of 0.98 (SE = 0.01) and 0.91 (SE = 0.03) on 666_N and 3000_V, respectively, and so were LIN and JCN with $\rho = 0.97$ and $\rho = 0.94$. Additionally, in comparison with human judgements, these measures, as a single group, also showed a similar pattern over the subsets of 666_N and 3000_V in Figure 5. Overall, the results on mutual proximity suggest that both the edge-counting and information content models may behave less distinctively in predicting semantic similarity, although multiple factors such as concept depth and frequency can be applied to differentiate the uniform distance. The simple edge-counting may prevail in calculating taxonomic similarity.

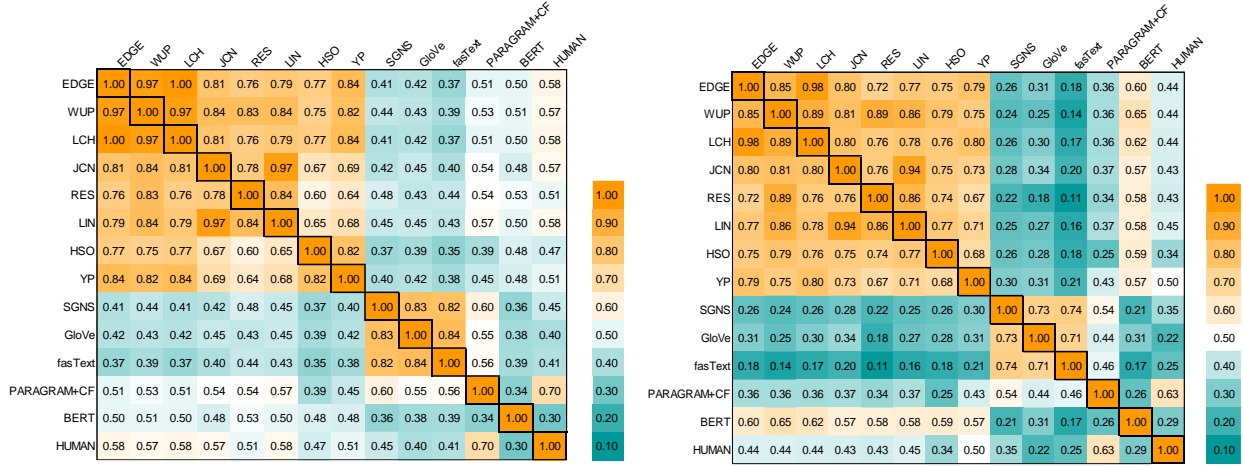

Figure 5: An overview of model proximity: 666_N (left) and 3000_V (right).

Our experiments are in line with previous findings on the effectiveness of concept specificity on taxonomic similarity measures (Jiang and Conrath 1997, Lastra-Díaz and Garcia-Serrano 2015), which may be strongly dependent on the organization of concept networks in mitigating the bias of uniform distance. Wang and Hirst (2011) showed that concept density alone held no correlations with human similarity judgements, in which the uneven concept distribution in WordNet could result in a lack of information inheritance, also claimed by Resnik (1995) in proposing information content. Concept density was then redefined through recursively adding information inheritance from a concept's ancestors (hypernyms in the IS-A hierarchy), and concept depth through designing a cumulative distribution function. They attained a moderate improvement on the similarity measures of Wu and Palmer (1994) and Jiang and Conrath (1997). Note that apart from the aforementioned concept depth and density in Section 3, other structural features in a taxonomy such as concept descendants (Seco et al. 2004), concept ancestors, and even leaf nodes (Sánchez et al. 2011) were also used to estimate concept specificity. Lastra-Díaz and Garcia-Serrano (2015) observed that merely minor differences existed while applying different strategies of intrinsic and extrinsic information content on the popular taxonomic similarity models in WordNet (Pedersen et al. 2004). These past studies have suggested that the taxonomy structures in LKBs may constrain different hypotheses of concept specificity from adjusting the uniform distance, and therefore the shortest path length may be the only effective factor in measuring taxonomic similarity.

Moreover, previous tests (McHale 1998, 2003), carried out for the adaptability of edge-counting, concurred with our initial results in the evaluation. Using Resnik's 28 noun pairs (1995), McHale (1998) tested the adaptability of the taxonomy similarity methods: EDGE, RES, and JCN (without the intervention of concept depth and local concept density) in WordNet and Roget. He found that EDGE significantly outperformed RES and JCN in Roget with $r = 0.89$, although it was the worst with $r = 0.67$ in WordNet. RES achieved $r = 0.79$ in WordNet and performed nearly the same in Roget. As for JCN, the best correlation value was $r = 0.83$ in

WordNet, which dropped to $r = 0.79$ in Roget. McHale claimed that the improvement of EDGE in Roget was partly due to its shallow structure—generally no more than eight levels; partly due to abundant lexical relationships (syntagmatic and paradigmatic). In contrast, WordNet is primarily equipped with the paradigmatic relationships of IS-A, and the 16 levels of IS-A hierarchy are relatively complicated. The moderately consistent results for RES, along with the slight variation of JCN in WordNet and Roget, indicated that they might be insensitive to the shortest path length in edge-counting, but might strongly depend on word usage statistics in practice. Jarmasz and Szpakowicz (2003) also investigated the adaptability of EDGE in WordNet and Roget, along with other methods implemented in the similarity package of Pedersen et al. (2004). Their evaluation results on the 30_N (Miller and Charles 1991) and 65_N partly fitted with McHale: for 30_N, EDGE outperformed others with $\rho = 0.87$ in Roget, and arrived at moderate correlation with $\rho = 0.76$ in WordNet; for 65_N, it performed nearly identically in Roget ($\rho = 0.81$) and WordNet ($\rho = 0.80$). The findings on the adaptability of the taxonomic similarity models in WordNet and Roget demonstrated that the shortest path length alone in the simple edge-counting can predict word similarity as effectively as the other complicated measures that consider fine-tuning the uniform distance with concept specificity. Irrespective of the idiosyncratic organization of LKBs, the shortest path length can be a prime factor in modelling taxonomic similarity.

### 6.2. *Distributional similarity in neural embeddings*

In Figure 5, the neural embeddings, as a single group, demonstrated little interrelations with the taxonomic similarity measures on 666_N (mean 0.45, 95% CI 0.43 to 0.47) and 3000_V (mean 0.33, 95% CI 0.29 to 0.38), in which SGNS, GloVe, and fastText were highly approximate to each other on 666_N (mean±SE: 0.83±0.01) and 3000_V (0.73±0.01). Similar results for the NNEs can also be found in Figure 4, in comparing with human similarity judgements on 666_N and 3000_V. The results showed that although different corpora and training methodologies were adopted in deriving distributional representations, SGNS, GloVe, and fastText may bear a close resemblance in computing distributional similarity, which confirms the previous hypothesis that they function equivalently as a process of matrix factorization (Levy and Goldberg 2014b, Levy et al. 2015). For example, in utilizing word co-occurrence matrix to yield neural embeddings, it has been demonstrated that SGNS works merely with factorizing a PMI transferred matrix (Levy and Goldberg 2014b), and GloVe's training objective is just to learn a factorization of a logarithmic matrix approximately (Levy et al. 2015).

#### 6.2.1 *PARAGRAM+CF*

PARAGRAM+CF was less associated with the unified embeddings of SGNS, GloVe, and fastText on 666_N (0.57±0.01) and 3000_V (0.48±0.03), together with BERT (0.34 and 0.26). The results of PARAGRAM+CF in Figure 4 and 5 also imply that through retrofitting neural embeddings with multiple semantic relationships, PARAGRAM+CF can reconstruct a novel latent semantic space that is probably distinct from the unified and contextualized NNEs and can substantially improve distributional vector models in similarity prediction (Mrkšić et al. 2016).

#### 6.2.2 *BERT*

As for BERT in Figure 5, although it was weakly related with the group of SGNS, GloVe, and fastText on 666_N (0.38±0.01) and 3000_V (0.23±0.03), it showed moderate association with the taxonomic similarity measures (0.50±0.01 and 0.60±0.01), better than the unified embeddings (0.41±0.01 and 0.24±0.01) and PARAGRAM+CF (0.50±0.02 and 0.36±0.02). The results partly indicated that a varied degree of architecture complexity on homogeneous neural networks might influence NNEs in deriving distributional similarity.

Ethayarajh (2019) and Jawahar et al. (2019) also found that BERT can generate hierarchical features from the lower layers of morphological information to the upper layers of syntactic and semantic information. Furthermore, Bommasani et al. (2020) investigated how to yield the static or unified embedding from each layer of NLMs such as BERT and GPT-2. Different from our method of using a synset's gloss in WordNet to generate sense or contextualized embedding, they first retrieved a word's contextual sentences from corpora, which were then re-injected into NLMs to generate an aggregated word embedding. With the increasing number of contextual sentences, they found that word embeddings derived from each layer can be gradually enhanced on correlating with human similarity judgements, outperforming the unified NNEs: SGNS and GloVe. To generate embeddings from BERT, we only employed a synset's gloss as its contextual sentence, whereas Bommasani et al. (2020) used at least 50 contextual sentences for each word, i.e. collecting 100k sentences in total for 2,005 target words. Given that nouns and verbs in WordNet have 2.79 and 3.57 senses on average, we input much fewer number of contextual sentences than Bommasani et al. (2020) in yielding neural embeddings, which may be the reason for the inferior performance of BERT in Figure 4. However, our gloss-based method can be helpful in WSD because we maximized sense similarity on the contextualized NNEs to harvest word similarity, which may be beneficial for disambiguating word meanings in context. Note that since our main task is to study the difference between taxonomic similarity and three types of NNEs, it is beyond the scope of this paper to investigate each layer of BERT in generating sense or contextualized embeddings, and we only examined BERT from the $9^{th}$ to $12^{th}$ layer, together with layer pooling.

## 7. Conclusion

After evaluating the popular taxonomic similarity methods, we can delineate three factors in estimating semantic similarity:

- **The shortest path length,** as a prime factor, can effectively measure semantic distance in edge-counting. Alternatively, information content only employs the shortest path to locate a *ncn* rather than its length for estimating concept specificity.

- **The link type** factor is mainly concerned with how to traverse a hierarchy. The link type is associated with the observation that concept proximity within the same distance is not universally applicable and may be fine-tuned using either the infrastructures of semantic networks such as concept depth and density or the concept usage statistics such as the IC of concept linkage.

- **The path type** factor only applies when different hierarchal relationships such as IS-A and HAS-A are merged into one similarity model. The reason is that concept similarity could be miscalculated on the hypothesis of uniform distance in heterogeneous hierarchies. With the fusion of the different hierarchies into one similarity model, the path type could improve the estimation of word similarity in a poly-hierarchy semantic network such as WordNet.

The findings of this study suggest that the shortest path length in edge-counting can hold a primary role in modelling taxonomic similarity, but its effectiveness may highly depend on the infrastructures of LKBs. The advantages of edge-counting models clearly lie in the fact that they can work on all comparisons of concept relationships, irrespective of concept metaphor, literalness, or usage statistics. With the introduction of concept usage in fine-tuning the uniform distance, calculating taxonomic similarity through information content may result in the exclusive use of the predominant senses given the skewed distribution of word senses in corpora.

Our experiments also confirm previous findings in the literature that taxonomic similarity can effectively and efficiently yield semantic similarity, outperforming distributional similarity on NNEs. With the fusion of neural embeddings and hand-crafted concept relationships, PARAGRAM+CF can substantially improve distributional semantic models in similarity prediction. Retrofitting concept relationships in LKBs into neural embeddings may provide a feasible way of combining both lexical knowledge and word usage patterns for similarity prediction, and the knowledge-augmented NNEs may further benefit other deep NLP tasks through transfer learning.

Owning to the substantial performance disparity between each measure and human similarity judgements, further studies are still needed to investigate how to adapt taxonomic and distributional similarity models under the limits of word frequency, polysemy degree, and similarity intensity. Taxonomic similarity may be improved using LKBs that contain multiple semantic relationships, e.g. a BabelNet of integrating both WordNet and Wikipedia. Apart from boosting distributional vector models through data augmentation such as increasing corpora size, controlling document length, and document domains, neural embeddings can be fine-tuned in many ways such as adjusting the size of a sliding window, optimizing hyperparameters in the stochastic gradient updates, and decreasing over-parameterization (Rogers et al. 2020). Various fusion methods are also worth investigating for harvesting advantages of knowledge resources and NNEs. Moreover, in contrast to the unified and hybrid NNEs, further studies on NLMs may focus on how to combine different static embeddings from each layer to enhance the quality of CNNEs in similarity prediction. As this paper aimed to conduct a comparative study of the typical WordNet-based taxonomic and NNEs-based distributional similarity models, there is some likelihood that discrepancies from our study would have arisen if more such measures or datasets had been embraced. Further studies on the current topic are therefore recommended to establish a hybrid similarity model of leveraging both taxonomic and distributional semantic features.

## Acknowledgments

This work has been supported by National Social Science Foundation (China) (17BYY119). We would like to thank the anonymous reviewers for their constructive comments and suggestions.